# Current Trends and Approaches in Synonyms Extraction: Potential Adaptation to Arabic

Eman Naser-Karajah
*Palestine Polytechnic University*
198005@ppu.edu.ps

Nabil Arman
*Palestine Polytechnic University*
narman@ppu.edu

Mustafa Jarrar
*Birzeit University*
mjarrar@birzeit.edu

***Abstract*— Extracting synonyms from dictionaries or corpora is gaining a special attention as synonyms play an important role in improving NLP application performance. This paper presents a survey of the different approaches and trends used in automatically extracting the synonyms. These approaches can be divided into four main categories. The first approach is to find the Synonyms using a translation graph, The second approach is to discover new transition pairs such as (Arabic – English) (English – France) then (Arabic-France). The third approach is to construct new WordNets by exploring synonymy graphs, and the fourth approach is to find similar words from corpora using Deelp Learning methods, such as word embeddings and recently BERT models. The papers also presents comparative analysis between these approoaches, and highlights potential adaptation to generate synonyms automatically in the Arabic language as future work.**

## I. INTRODUCTION

Synonyms are important in many application areas in Natural Language Processing (NLP) [34]. In Information Retrival, synonyms can use to expand queries and retrive richer results. It could also be beneficial in automatic text summarization, which is employed to identify repetitive information to avoid redundant summaries. In language generation, synonyms are used to create more varied texts [13].

Two words are considered synonymous if they can be interchanged in the same sentence without changing its meaning, such as {arrange, organize}. Synonyms are defined in [2] as "*two expressions are synonymous if the substitution of one for the other never changes the truth value of a sentence in which the substitution is made*." A more formal definition of synonymy, in ontology engineering, is *"a formal equivalence relation (i.e., reflexive, symmetric, and transitive)."* thus "*Two terms are synonyms iff they have the exact same concept (i.e., refer, intentionally, to the same set of instances). Thus,* $T_1 =C_i T_2$*. In other words, given two terms* $T_1$ *and* $T_2$ *lexicalizing concepts* $C_1$ *and* $C_2$*, respectively, then* $T_1$ *and* $T_2$ *are considered to be synonyms iff* $C_1 = C_2$*"* [12].

Synonyms can be automatically extracted from different sources and in different ways, such as dictionaries, WordNets, or language models like Word Embedding [33]. Synonyms could be extracted for a general language such as or domain-specific such as a medical field [6].

One of the known synonymy databases for English is the Princeton WordNet (PWN) [2], which was built manually at Princeton University as a network of lexical concepts. Sets of synounyms (called Synsets) in WordNet are synonyms connected by semantic relations such as hypernyms and meronyms. Although there are many worndents that we built of other languages, following the same way PWN was built, but most of them are small. This is because manual construction of wordnet is time-consuming and expensive.

Synonyms extracted from statistical language models, like Word2Vec embeddings, can be defined as "closely-related" words [7]. scuh language models capture context similarity between words by converting the words into vectors. Words are considered similar if their vectros are close to each other – typically measured using cosine similarity. As Word Embeddings are static models (i.e., does not represent the meanings in different contexts), contextualized embedding models, like BERT, are recently used to extract synonyms, showing and proving their power for generating synsets in dynamic contexts. These models use lexical substitution, which aims to replace the target word with the most suitable synonyms without changing the sentence's meaning [24].

This paper surveys the most relevant works related to extracting synonyms, focusing on the extraction techniques and their evaluation methods and datasets. We cluster these works into four groups which will be presented as follows: section 2 presents approaches to find synonyms using translation graphs. Section 3 presents approaches to discover new transition pairs. Section 4 describes the approaches that construct new WordNets approaches by exploring synonymy graphs. Section 5 discusses approaches based on Word Embeddings. Section 6 presents a comparative analysis for all the discussed techniques. Section 7 presents directions for future work, and finally, section 8 presents the conclusion

## II. EXTRACTING SYNONYMY USING TRANSLATION GRAPHS

Some researchers proposed to build synonyms using translation graphs, which can be language-dependent or independent. The idea is to take bilingual dictionaries as input to build a translation graph between two words in a given language, find the translations for it in the other language, and then extract the paths which present synonyms. The key idea here is how to extract these paths. The accuracy of the generated synonyms depends on the structure and the accuracy of the input dictionaries. A recent approach [6] suggests to convert a bilingual dictionary into an undirected translation graph. The approach has two steps. The first step is to find all possible sets of candidate synonyms from cyclic paths. The propagation of the translation stops when it reaches the root word (i.e., cycle), when no translations are found, or reached the maximum number of *k* levels. Figure 1 presents an example of a translation graph for the words (غَابَة) as in [6]. If a path was found ($\boldsymbol{a1} \rightarrow e1 \rightarrow a2 \rightarrow e2 \rightarrow a3 \rightarrow e3 \rightarrow \boldsymbol{a1}$) then all words in the cycles are converted into sets of bilingual synonyms such as {a1, a2, a3} = {e1, e2, e3}. The second step is to consolidate the candidate synsets with the same translation by taking the union between them. To evaluate this approach, the authors converted the bilingual synsets found in the Arabic WordNet (AWN) into a flat bilingual dictionary then rebuild the AWN again using their algorithm. The accuracy is measured by cosine similarity between the original AWN and the generated AWN; the accuracy reached 82%. The algorithm results depend on the accuracy of the bilingual input dictionaries.

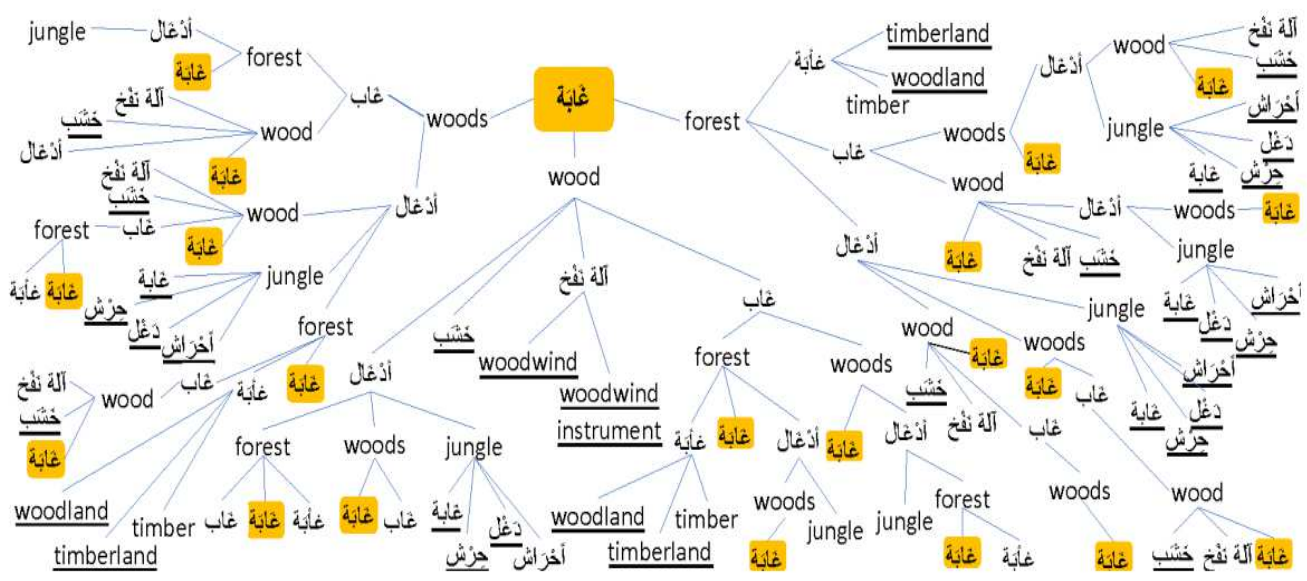

Fig.1: Translation graph

A similar approach was used to enrich an English-Italian thesaurus's quality by discovering missing synonyms [8]. The authors claimed that their approach could be generalized and used for other languages; it can also improve the user dictionary's quality. They reach up to 80% accuracy at the sixth level. Their approach takes each word in the thesaurus and finds all corresponding translations. A directed translation graph Cyclic and Quasi-Cyclic (CQC) are constructed as illustrated in figure 2 [8]. Depth First Search discovers the paths using the scoring function, which weights paths based on path length, the shorter path takes higher weight, then rank these paths. The algorithm is evaluated using the TOFEL dataset and compares their results with other algorithms used in the same datasets.

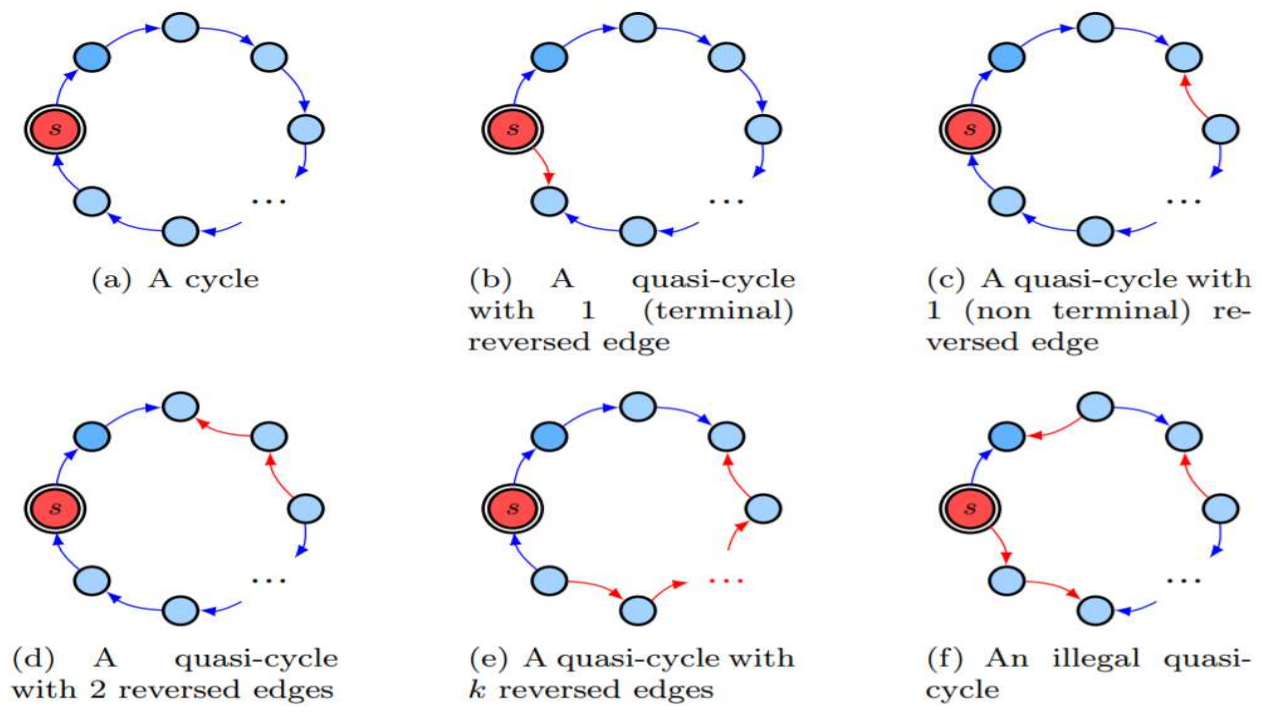

Fig.2: Cycles and quasi-cycles construction

The main difference between the two approaches, [6] and [8] is how the paths are extracted. No Quasi-Cycles are used in [6], and more importantly a consladation phase is introduced instead of using phath weight as in [8]. Nevetheless, both approaches build translation graphs from bilingual dictionaries inorder to extract synonyms. Although this method's precision is promissiing, their results depend on the accuracy of input dictionaries.

## III. Discovering new transition pairs

Other researchers proposed to find new translation pairs using translation graphs, which a task that is related to extracting synonyms. The discovered pairs offer new translations pairs between languages that haven't bilingual dictionaries for translation between them. It can also be used to enrich the existing bilingual dictionaries with newly discovered translation pairs.

A recent approach presented in [9] presents three algorithms for the automatic discovery of new translation pairs between languages based on bilingual input dictionaries. The cycle-based algorithm is used to build the translation graphs between the selected bilingual dictionaries in Apertium dictionaries to find translations between Portuguese, French, and English with a cycle length of at least 4. The path-based algorithm assigns weight to each path in the translation graph-based translation pairs number in each path. Paths with a short length and higher frequency take lower weights. The multi-way neural machine translation is used to generate the target pairs' translations. They used parallel corpora for the selected languages (Spanish, Italian, French, Portuguese, Romanian, and English) as the translation pairs are English-Spanish, French-Romanian, and Italian-Portuguese. To generate the target translation, they add the output for cycle and path algorithms. They evaluate their algorithm by randomly extracting translation pairs and comparing them with manual pair translations for the same selected pairs. Their approaches show very low recall and an acceptable precision (25-75%).

Another approach was proposed in [10] to discover new translation pairs from Apertium RDF Graph, containing 22 Apertium bilingual dictionaries. The translation graph is built by building context for each word. The context is computed by finding all translations for each word of three levels, then calculating cycles occurring in context. They restricted the process to be up to level 7. The dense cycle is calculated to find the confidence score to accept word translation. Two experiments were used to evaluate this approach. The first one is done by removing the English-Spanish, Spanish– English from the Apertium RDF graph, then extract it again using their algorithm to compare the two results. The algorithm was able to extract 72% of the original translations. The second experiment is done by finding a new English-French translation, which does not exist in the Apertium RDF graph. The algorithm was able to extract 73% of the translations in the converted wiktionary English-French file compared to the converted wiktionary English-French file.

In these kinds of approaches, extracted new transalation depeneds on the number of translations for the target language, in case it is higher, a higher recall can be expected. In addition, wrong translation may occur due to polysemy. The problem can be reduced by using many languages to generate correct new translation pairs. For example, an OTIC algorithm was built in [25] to detect the wrong translation in creating a bilingual dictionary using another language. The approach is extended in [26] to develop multilingual lexicons from bilingual lexicons..

## IV. Constructing new WordNets

This section presents related approaches to construct WordNets automatically focusing on synonymy extraction.

One of the most critical parts for the automatic construction of synonyms is evaluating the accuracy, precision, and recall, as there is no standard evaluation methodology and dataset for that. Lexical databases or word embedding-based approaches are currently being evaluated by comparing with existing WordNets or manually evaluating the generated synsets [3]. In [11], the authors proposed constructing a translation graph from multiple Wiktionaries to create word pairs with the same meaning for at least one synset. Each synset would have a complete subgraph when the translations between all terms with the same meaning are constructed in the undirected graph. Figure 3, from [11] illustrates synsets extraction and expansion using three algorithms, clustering, greedy, and supervised learning presented. The clustering algorithm is used to extract the synsets; the input WordNet is used to map the discovered synsets by clusters to PWN. The Greedy algorithm is used to extend the synsets with two metrics, belongingness and coverage, which are used to assign scores to word synsets pairs; the input WordNet is used to initialize the synsets and expand. The supervised binary classification is used to classify the new words which are included in synsets. The translation graph is linked with the existing WordNet for the selected language to induce new synsets using a clustering algorithm. The approach is evaluated by building Slovenian, Persian, German, and Russian WordNets from scratch and comparing the results with original WordNets. The accuracy varies from 20% up to 88%. Generating synonyms using machine learning algorithms such as Greedy algorithms [11] has any parameters and features that could affect algorithms results, such as choosing the threshold values. It must be set using a validation set that is separate from the test set. The generated synsets will be smaller but accurate for high threshold values when comparing them with WordNet synsets. It can reach above 90% accuracy. A lower threshold value generates larger synsets. But it should be followed by a filtration step using a more detailed monolingual corpus.

A language-dependent approach was presented in [13]. This approach uses three kinds of input data: bilingual English-Chines corpus, Chinese monolingual corpus, and Chinese monolingual dictionaries, which are used to extract synonyms automatically and separately for each input, then results for each input are combined. For the monolingual dictionary, the hubs and authorities are used and considered as word features, then a feature vector for each word is constructed. The synonyms are extracted from the Chinese monolingual dictionary by measuring the cosine similarity between two words using the feature vectors. In the bilingual carpus, the synonyms are extracted using the word translation using the English-Chinese bilingual dictionary, then assign them a translation probability based on English–Chinese corpus. The Chinese monolingual corpus is used to find the words in the same contexts by finding their named attributes' dependency triples. For each word in the carpus, they find the relation of the word with the next ones (word1, relation type, word2); if these attributes are the same, they are considered synonyms. The results of the three algorithms are combined using a binary classifier. This approach is evaluated by comparing the compound synonyms from the tree algorithms to the compound WordNet and thesaurus synonyms. WordNet construction from monolingual corpora uses considerable corpora resources in machine translation or Word Embedding. As they provided a vast vocabulary and context information, these approaches recall is usually higher than multilingual graph-based approaches. Using more than one input type to enrich the graph is not always needed to get good results. Building larger translation graphs by integrating other bilingual dictionaries may improve the results.

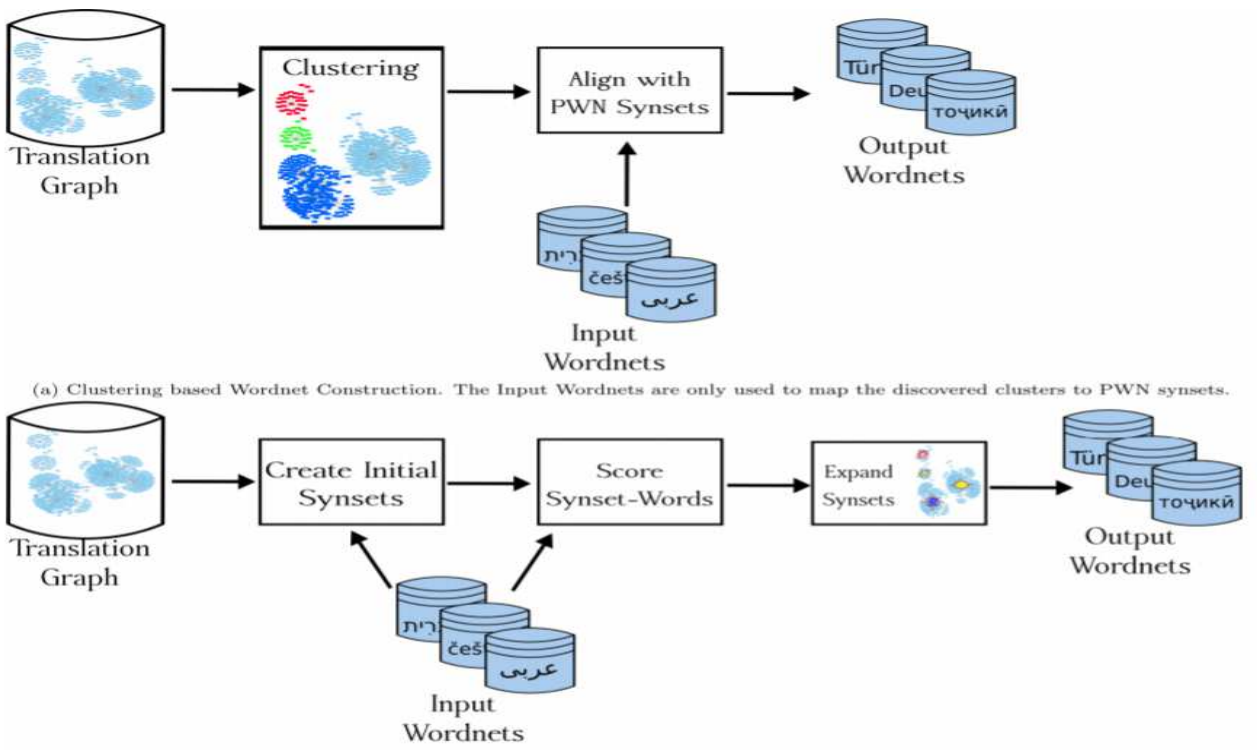

Fig.3: Synsets extraction and expansion

Another effort in [14] is to generate new WordNet synsets using existing WordNets, machine translations, and bilingual dictionaries. The authors considered their approach as language-independent. Each word in the PWN synset is translated using machine translation to the target language. Then they check for irrelevant translations because of polysemy using intermediate WordNet or bilingual dictionary of the target language. A rank method for the translated synsets is then applied. The higher rank implies a word belonging to the new WordNets corresponding in the target language. This approach was evaluated manually by volunteers who use the WordNet's language as their mother tongue; their judgment is given as rating quality from 1 (Poor) - 5 (excellent). They randomly select 500 synsets and compare the volunteer judgment. They achieved an average score of 3.78/5.00 or 75.60%.

Other related approaches include the use of senses in Wiktionary [28], finding missing edges and consolidating similar synsets [29], finding semantic relations between concepts using wordnets and Wikipedia [30], building a graph of relationships between nodes using bilingual dictionaries, monolingual corpora, mono/multilingual

Thesauri, and Wiktionary [31], as well as using an ontology, Bablenet, and Google Translate [32].

## V. EXTRACTING SYNOYNOMS USING LANGUAGE MODELS

This section presents approaches to generate synonyms using static and dynamic Word Embedding.

Recently, neural network models are used for extracting synonyms using a Word Embedding model from corpora. [7] presented an approach for extracring Arabic synsets. It can be considered a general approach with a two-step algorithm. In the first step, words in the NewsCrawl 2014 corpus is tagged with POS; then, Word Embedding is constructed by converting every word into vector presentation. Only the words with a frequency of 25 are considered. The skip-gram model is used to group similar words —the cosine similarity is then used to cluster the words into adjectives, nouns, adverbs, and verbs. In the second step, the approach is evaluated by the Simlex-99 dataset, a forward neural network, and a Word Embedding list from the first step to discover the synonyms. They achieve 76% accuracy.

Another approach in [15] uses Word Embedding to enhance wordnets. This approach is an extention to the approach presented in [14]. After a translation graph for the target language is constructed from PWN, Word Embeddings are used to improve the translation step's synsets quality. Irrelevant words were removed in the synsets by calculating the cosine similarity between the words in the synset. A synset similarity threshold value is used to accept or dismiss the word from the synset. The same is done to validate the synsets relations. The cosine similarity is calculated between all synset pairs. When the values are above the threshold, then it is maintained otherwise discarded. They evaluated their approach by generating Arabic WordNet AWN using the Wattan-2004 corpus. Then they create the Word Embedding using a continuous bag of words (CBOW). They generate 60% of the manual AWN with a precision of 78.4%.

A recent approach is proposed in [16] to construct WordNets using Word Embeddings and machine translations. A translation of a target word using a bilingual dictionary or machine translation is extracted, then used to find nominee synsets from the PWN. After that a score to each synset assigned to each synset. Ranking of the resulted synsets is done by computing cosine similarity with Word Embeddings. The experiment is done for nouns, adjectives, verbs, and adverbs, and their results were promising; they get 90-94%. The main challenge is that the resources of all language pairs are required, which is not easy to find.

Another an unsupervised learning methodology is discussed in [20]. This approach consists of two steps; the first step is to capture the best settings for finding synonym relations by training the Word Embedding using Word2Vec, CBOW, and SG models by two corpora, the KSUCCA and Gigaword. The preprocessing step includes tokenization, diacritic, numbers, English letter removals, and normalization. The next step is to filter the similar emending list to catch the synonyms using the synoExtactor pipeline. This pipeline firstly finds the most similar words by calculating the cosine similarity for preselected words from the two corpora; the high score implies the same context, which contains synonyms and the other relations such as anatomy. Secondly, they apply three filters, the Lemmatization filter, which removes inflections; they create a dictionary of lemmas, for each lemma, all words with the same lemma from carpus are added. The second filter is the collection filter that uses the collocation dictionary generated from corpora. The stop words and the collection words that have appeared less than five times were removed. The last filter is the POS filter which keeps the words with the same POS; synonyms always have the same POS. Figure 5 presents the SynoExtractor pipeline [20]. The approach is evaluated by comparing the generated synsets with the Ama'any Arabic thesauri; a manual evaluation by two linguistics is also done. The SynoExtractor reached 60% precision for the KSUCCA corpora and 74% precision for Gigaword corpora.

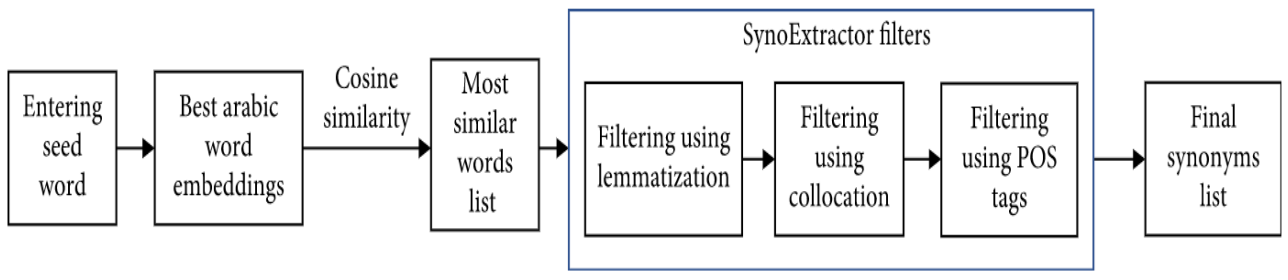

Fig.5: SynoExtractor pipeline

As presented in the Word Embedding approaches [7,15,16,20], cosine similarity is not enough. It may extract the most similar list of words representing synonyms and extract other relations such as inflections and antonyms.

Another approach in [21] developed an automatic extraction model of synonyms to create Quranic Arabic WordNet (QAWN) by the use of three resources: the boundary annotated Qur'an, lexical recourses for collecting derived words from the Qur'an, traditional dictionaries. In the first step, they present the Holy Qur'an using Vector Space Model (VSM). In the second step, using the traditional Arabic dictionaries, they extracted the Qur'an words meaning. They used term frequency and inverse document frequency in the vector space model and compute cosine similarities between Quranic words depending on textual definitions extracted from traditional Arabic dictionaries. The words with the highest similarity were clustered to form a synset. They evaluated their approach using an information retrieval system. They reached 34.13 % recall and very low precision as clustering words could not generate actual synsets as they contained other word relations.

In Word Embedding approaches, supervised techniques help generate or filter more accurate synsets when comparing unsupervised techniques. Supervised methods need a labeled data used to nominee the synonyms from other relations, not for extracting relations. While the unsupervised techniques didn't need a labeled data, they can use any raw carpus using the clustering techniques. Still, it needs another approach to filter the synonyms from other relations

Another novel approach is presented in [22]. It uses the BERT-based lexical substitution model, which is inspired from [24]. The model extracts the relevant synsets within the same contexts and extracts the most suitable synonyms for the target word without changing the sentence's meaning by measuring the contextual representation similarity. The novel approach uses a partial drop masking for the Embedding of the target word by setting it to zero then predicts the target word based on its position on the sentence by calculating the validations score. By ranking the validation scores, the higher values present the target word's synonyms. It extracts the most suitable synonyms for the target word without changing the sentence's meaning. This approach is evaluated using two benchmarks datasets; the SemEval 2007 dataset (LS07) and the CoinCo dataset (LS14) using five metics ( best, best-mode, out-of-ten, out-of-ten-mode, and Precision @1 ). The metrics that present the best predictions' quality are evaluated using best, best-mode, and Precision @1. While the coverage of the synsets is evaluated using out-of-ten and out-of-ten-mode. The results show that their embedding dropout to BERT performs better than the previous works on the same two benchmarks datasets. They reached 51.1 % and 56.3 % in Precision@1 for the LS07 and LS14 datasets, respectively.

This approach permits the partial BERT to have balanced consideration for the target word's semantics, which will generate only the relevant synsets for the sentence meaning.

BERT is a Word Embedding approach that can either mask the Embedding of the target word or not. In masking the Embedding of the target word case, the BERT generates synonyms for the target word, which do not fit even in the same context. While in the second case, which doesn't mask the Embedding of the target word, it will generate the same word as a synonym. The approach presented in [22] makes partial masking for the target Word Embedding, then predicts the target word based on its position on the sentence by calculating the validations score. This approach extracts only the most suitable synonyms for the target word without changing the sentence's meaning, which is proven through their precision results that exceed the older approaches' precision values. The main difference between the BERT approach and the partial masking BERT approach discussed in [22] is that the BERT approach may ignore suitable synonyms for the target word. Besides, they don't consider the sentence meaning. In contrast, the partial BERT has balanced consideration for the target word's semantics, which will generate only the relevant synsets for the sentence meaning.

## VI. Comparative analysis

This section presents a comparison between the different approaches presented in the previous sections. The paper's contribution includes presenting current techniques and approaches for automatic synonyms generations and their evaluations and standard guidelines for building and improving the synset generation. Table 1 presents a comparison between the different approaches, which are ordered ascending by year. The comparison is based on the input resource, the methodology used, the evaluation methods used to validate the approach results, and the approach accuracy. Even though the paper is categorized into four main approaches, another categorization can be done depending on the input resources to generate the synsets: lexicon base and Word Embedding base [3,33].

Some of the four group approaches are language-dependent, which is done for a particular language and cannot extract synonyms for other languages. While most approaches are language-independent; they can be used as a general methodology to extract synonyms in other languages; this depends on the methodology used and the approach algorithm.

## VII. Future directions

The presented approaches are to be investigated for adaptation to generate synonyms in the Arabic language automatically. The four categories which are discussed above can be implemented on the same Arabic dataset [18, 37,38]. The evaluation results can be compared to determine the best approach to generate Arabic language synonyms with the highest accuracy. The highest accuracy implies the best approach to be applied as the Arabic language needs special attention in synonyms generation.

A vast manual Arabic- Multilingual database for Arabic lexicon in [18,35,36] contains 150 Arabic–multilingual resources that can be used, such as modern and classical dictionaries, AWN, and thesauri. The best approach accuracy for the Arabic language can be used to generate Arabic synonyms. In other words, it can be used to build a new AWN. The existed AWN contains 10,000 Arabic synsets, which is relatively small. The new AWN could be mapped to generate Arabic synsets equivalent to the PWN containing 117,000 English synsets.

## VIII. Conclusion

This paper presents a review of several approaches for automatic synonyms generations. The paper divided these approaches into four types: synonymy extraction using translation graphs, discovering new transition pairs, constructing new WordNets, and extracting synonyms using Word Embeddings. Automatic Synonyms generations is a promising research area, as generating manual synonyms is costly and time-consuming.

For the Arabic language, adding POS, diacritics, or other morphological features can lead to more accurate synonyms but need more attention. The extracted synonyms could be domain-specific or general as Arabic WordNet. The presented approaches could be investigated and evaluated thoroughly for potential adaptation to automatically generate synonyms in the Arabic language.

| Num | Approaches | Paper name | Year | Methodology | Language | Input | Evaluation | Accuracy |
|---|---|---|---|---|---|---|---|---|
| **1** | **Extracting Synonymy using translation graphs** | Cycling in graphs to semantically enrich and enhance a bilingual dictionary [8] | 2012 | Builds a directed translation graph (cyclic and quasi-cyclic) | A general approach | 1. English-Italian<br>2. Bilingual dictionary | Use the TOFEL dataset and compares their results with other algorithms used in the same datasets. | 80% |
| | | Extracting Synonyms from Bilingual Dictionaries [6] | 2021 | 1. Builds a translation graph<br>2. Synsets consolidation | A general approach | Arabic-English Bilingual dictionary | Rebuild the AWN again using their algorithm | 82% |
| **2** | **Discovering new transition pairs** | Leveraging RDF graphs for crossing multiple bilingual dictionaries [10]. | 2016 | Aperteum RDF Graph | A general approach | 22- Aperteum RDF Graph<br>1. English- Spanish<br>2. English - France | 1. Extract English-Spanish, Spanish–English using their algorithm.<br>2. Find a new English-French translation, which does not exist in the Apertium RDF graph | English - Spanish 72%<br>English – France 73% |
| | | Leveraging Knowledge Graphs with Neural Machine Translation for Automatic Multilingual Dictionary Generation [9] | 2019 | 1. Graph-based.<br>2. Path-based<br>3. Multi-way neural machine translation | It can't be general – language-dependent | 1. Bilingual dictionaries English -Portuguese<br>2. Carpus | Choose randomly extracted translation pairs and comparing them with manual pair translations for the same selected pairs | 25-75% |
| **3** | **Constructing new WordNets** | Optimizing synonym extraction using monolingual and bilingual resources [13] | 2003 | 1. The synonyms extracted from Chinese monolingual dictionaries<br>2. Word translation using Bilingual carpus (English –Chinese)<br>3. Chinese monolingual corpus to find words in the same context | It can't be general – language-dependent | 1. Bilingual carpus (English –Chinese)<br>2. Chinese monolingual corpus<br>3. Chinese monolingual dictionaries | Compare the compound synonyms from the tree algorithms to the compound WordNet and thesaurus synonyms. | low accuracy |
| | | Automatically constructing WordNet synsets [14] | 2014 | 1. Machine translation<br>2. Ranking translated synsets | A general approach | 1. Existed WordNet<br>2. Bilingual dictionaries | Randomly select 500 synsets and compare the volunteer judgment | 75% |
| | | Synset expansion on translation graph for automatic WordNet construction [11] | 2019 | 1. Clustering algorithm<br>2. Greedy algorithm<br>3. Supervised learning | A general approach | 1. Wiktionary<br>2. Existed WordNet | Built Slovenian, Persian, German, and Russian WordNets from scratch, then compare the results with original WordNets of these languages | 88% |
| **4** | **Extracting Synoynoms using Word embeddings** | Enhancing automatic WordNet construction using word embeddings [15] | 2016 | Word Embedding's | A general approach | 1. WordNet<br>2. Textual corpora | Generate Arabic WordNet AWN using Wattan 2004 corpora. Then they create the word Embedding using a continuous bag of words (CBOW). | 78% |
| | | Towards an automatic extraction of synonyms for Quranic Arabic WordNet [21] | 2016 | Use the term frequency and inverse document frequency Tf-idf in the vector space model (VSM) | It can't be general – language-dependent | 1. The Holy Qur'an,<br>2. lexical recourses for the Qur'an<br>3. Traditional dictionaries | An information retrieval system | 34.13% |
| | | Automated WordNet construction using word embeddings. [16] | 2017 | 1. Unsupervised learning<br>2. Machine translations<br>3. Word embeddings. | A general approach | 1. Existed WordNet PWN<br>2. French and Russian | Build 200 French and Russian test WordNets using Google translator as a machine translator, word embedding dataset in [17]. | 90-94% |
| | | BERT-based Lexical Substitution [22] | 2019 | Partially mask BERT | A general approach | LS07 trial | 1. SemEval 2007 ( LS07).<br>2. CoinCo (LS14) | 51.1 % for LS07<br>56.3 % for LS14 |
| | | Extracting Word Synonyms from Text using Neural Approaches [7] | 2020 | Two-step approach using NN to:<br>1. Build word embedding list<br>2. Use Word Embedding to train NN | A general approach | 1. Textual corpora<br>2. Sim Lex-999 lexicon | Use Simlex-99 dataset, a forward neural network, and a word embedding list from the first step to discover the synonyms. | 76% |
| | | A Novel Pipeline for Arabic Synonym Extraction Using Word2Vec Word Embeddings [20] | 2021 | Unsupervised with 2 step approach :<br>1. Build word embedding list from two corpora.<br>2. SynoExtractor pipeline to filter synsets | A general approach | 1. KSUCCA<br>2. Gigaword | 1. Compare the generated synsets with the Ama'any Arabic thesauri.<br>2. A manual evaluation by two linguistics | 60% for KSUCCA<br>74% for Gigaword |

Table 1: A comparison between the different synonyms extraction approaches